\documentclass[letterpaper, 10 pt, conference]{ieeeconf}  

\IEEEoverridecommandlockouts                              
\usepackage[utf8]{inputenc}
\usepackage[T1]{fontenc}
\usepackage{graphicx} 
\usepackage[hidelinks]{hyperref}

\title{\LARGE \bf
Fine‑Tuning BERT for Domain‑Specific Question Answering: Toward Educational NLP Resources at University Scale
}

\author{Aurélie Montfrond$^{1}$%
\thanks{*This work was based on my master's thesis}
\thanks{$^{1}$A. Montfrond, MSc in Artificial Intelligence,
Dept. of Electronic \& Computer Engineering, University of Limerick, Ireland
        {\tt\small 22289003@studentmail.ul.ie}}%
\thanks{}%
}

\begin{document}

\maketitle
\thispagestyle{empty}
\pagestyle{empty}

\begin{abstract}

Prior work on scientific question answering has largely emphasized chatbot‑style systems, with limited exploration of fine‑tuning foundation models for domain‑specific reasoning. In this study, we developed a chatbot for the University of Limerick’s Department of Electronic and Computer Engineering to provide course information to students. A custom dataset of 1,203 question–answer pairs in SQuAD format was constructed using the university book of modules, supplemented with manually and synthetically generated entries. We fine‑tuned BERT (Devlin et al., 2019) using PyTorch and evaluated performance with Exact Match and F1 scores. Results show that even modest fine‑tuning improves hypothesis framing and knowledge extraction, demonstrating the feasibility of adapting foundation models to educational domains. While domain‑specific BERT variants such as BioBERT and SciBERT exist for biomedical and scientific literature, no foundation model has yet been tailored to university course materials. Our work addresses this gap by showing that fine‑tuning BERT with academic Q\&A pairs yields effective results, highlighting the potential to scale towards the first domain‑specific QA model for universities and enabling autonomous educational knowledge systems.

\end{abstract}

\section{Introduction}

Universities rely heavily on their websites to provide prospective and current students with accurate information about courses and modules. However, navigating large volumes of content can be inefficient and time‑consuming. Chatbots offer a promising solution by enabling students to query course information in natural language and receive direct, context‑aware answers. Prior studies have explored university chatbots using rule‑based or hybrid approaches, demonstrating potential but limited scalability (Dinh \& Tran, 2023; Neupane et al., 2024). Recent advances in large language models have transformed the capabilities of such systems. Among these, BERT (Bidirectional Encoder Representations from Transformers) has become a benchmark for extractive question answering tasks due to its bidirectional transformer architecture (Devlin et al., 2019). Domain‑specific adaptations such as BioBERT, trained on biomedical corpora (Lee et al., 2020), and SciBERT, trained on scientific publications (Beltagy et al., 2019), demonstrate that fine‑tuning BERT on specialized datasets yields significant improvements in accuracy. Despite these successes, no foundation model has yet been fine‑tuned specifically for datasets focused on university course information. In this paper, we address this gap by fine‑tuning BERT using the Hugging Face Transformers library (Wolf et al., 2020) on a custom dataset of 1,203 question–answer pairs derived from the Department of Electronic and Computer Engineering in the University of Limerick’s book of modules. The dataset was formatted in SQuAD style (Rajpurkar et al., 2016) and included both manually authored and synthetically generated examples. We evaluate the fine‑tuned model using Exact Match (EM) and F1 scores, which are standard metrics for extractive QA. Our results show that even modest fine‑tuning improves knowledge extraction and reasoning in the educational domain. This work demonstrates the feasibility of adapting foundation models to university‑scale knowledge systems and highlights the potential for developing specialized educational models that can support students in accessing course information more efficiently.

\section{Related work}

Several studies have explored the use of chatbots in university contexts to improve access to course information and student services. Neupane et al. (2024) developed a chatbot for Mississippi State University using a Retrieval‑Augmented Generation (RAG) pipeline. By integrating private documents into the model, RAG improved accuracy compared to baseline LLMs such as ChatGPT. Their evaluation combined usability surveys (SUS) with performance metrics including precision and recall, and they highlighted the distinction between rule‑based and AI‑driven chatbot architectures. Hailu et al. (2024) focused on engineering department FAQs, comparing Support Vector Machines, Naive Bayes, and deep neural networks implemented with TensorFlow and Keras. Their results showed that deep learning achieved the highest accuracy (91.55\%), although the chatbot was trained in a non‑English language context. Usability was assessed through questionnaires with students and lecturers, and performance was measured using accuracy, precision, recall, and F1 scores. Cherumanal et al. (2023) built a chatbot for computer science programs at RMIT University, comparing intent‑based (IB) and RAG approaches. Their dataset was derived from FAQ documents, and evaluation included NDCG for retrieval effectiveness, ROUGE, and BERTScore to detect hallucinations. They found that IB models were more reliable for questions with no possible answers, while RAG models were better at synthesizing responses from multiple passages but more prone to hallucinations. Galstyan et al. (2024) created a chatbot for the American University of Armenia using RAG and vector databases, integrating OpenAI GPT‑3/4 and Meta Llama‑3 models. They emphasized user experience through a ChainLit interface and evaluated performance with CoLA and ROUGE metrics, finding GPT‑4 to be the most accurate but also the most costly. Attigeri et al. (2024) built a chatbot for the Manipal Institute of Technology, comparing five NLP models including neural networks, TF‑IDF vectorization, sequential modeling, and pattern matching. Their dataset was collected from social media and university sources, with preprocessing steps such as tokenisation, stemming, and lemmatisation. Sequential modeling achieved the best performance, preventing overfitting and outperforming other approaches. Okonkwo et al. (2021) reviewed 53 studies on chatbots in education, finding that most focused on teaching and learning rather than advisory systems for course information. They highlighted ethical concerns, including transparency and privacy, and noted challenges in evaluation, user attitudes, and NLP accuracy. Their findings emphasized the need for improved algorithms and regular data updates to maintain chatbot effectiveness. Vukomanović et al. (2022) implemented a chatbot for the Belgrade Business and Arts Academy of Applied Studies using intent/entity classifiers and Microsoft SQL Server. Their evaluation showed that the chatbot was most effective in attracting prospective students, with limited use by current students. Al Ghaithi et al. (2020) built a chatbot for Middle East College in Oman to address queries from existing and prospective students. Their system relied on a database of questions and answers, with administrative functionality to update entries and collect user feedback. While technically limited, the design emphasized usability and adaptability. De Bruyn et al. (2021) developed an FAQ chatbot using a retrieval architecture with embedding, transformer, and feedforward layers. Their dataset contained 1,200 questions mapped to 76 answers, and performance was measured using cross‑entropy loss and the Adam optimizer. Rashinkar et al. (2021) built a chatbot for Pune University using Naive Bayes classification within the Django framework and Chatterbot library. Their system matched questions to answers in a database, with unanswered queries forwarded to an admin for later inclusion. They compared Naive Bayes with Porter‑Stemmer and K‑means clustering, finding Naive Bayes to be the fastest and most effective. Together, these studies demonstrate the diversity of chatbot architectures applied in higher education, ranging from rule‑based systems to retrieval‑augmented generation and deep learning classifiers. However, none of these approaches directly investigate fine‑tuning foundation models such as BERT for extractive question answering on university course information. This gap motivates our work, which adapts BERT to the University of Limerick’s Department of Electronic and Computer Engineering dataset.

\section{Methods}

\subsection{Dataset preparation} We constructed a domain‑specific dataset from the Book of Modules of the University of Limerick, focusing on the Department of Electronic and Computer Engineering. Module descriptions and program web pages were extracted using web scraping techniques (Mitchell, 2018; Ryan, 2020) and manually organized into the SQuAD v1.0 format (Rajpurkar et al., 2016). The final dataset consisted of 1,203 question–answer pairs, created through a combination of manual annotation and large language model (LLM) assistance. Each generated entry was manually verified to ensure that answers matched the exact wording of the context, since extractive QA models require precise alignment to compute the \texttt{answer\_ start} index. A validation set was created by reserving 20\% of the dataset, with multiple answers per question added to support evaluation functions. The dataset was stored in JSON format, consistent with Hugging Face requirements (Wolf et al., 2020).

\subsection{Model selection} We fine‑tuned several BERT‑based architectures including some already fine-tuned on the original SQuAD datasets, as these models are optimized for extractive question answering:

\begin{itemize}

\item BERT large uncased (Devlin et al., 2019).
\item DistilBERT  (cased , uncased and distilled SQuAD) (Sanh et al., 2019).
\item RoBERTa base (SQuAD 2.0) (Liu et al., 2019).
\item BioBERT (base and large, v1.1 SQuAD) (Lee et al., 2020).
\item SciBERT (cased and uncased) (Beltagy et al., 2019).
\item XLM‑RoBERTa base (SQuAD 2.0) (Conneau et al., 2020).
\item QABERT small (SRDdev/QABERT-small, HuggingFace implementation).

\end{itemize}

These models were selected to compare performance across general‑purpose, biomedical and scientific variants of BERT‑based architectures.

\subsection{Preprocessing and tokenization}

The dataset was tokenized using Hugging Face’s \texttt{AutoTokenizer} (Wolf et al., 2020). Tokenization converts text into numerical representations that can be processed by transformer models. This involves two stages:

\begin{itemize}
    \item Splitting text into tokens — words or subwords depending on the tokenizer (Sennrich et al., 2016).
    \item Encoding tokens into numbers — producing \texttt{input\_ids} that represent each token numerically.
\end{itemize}

Each entry included \texttt{input\_ids}, \texttt{attention\_mask}, \texttt{start\_positions}, and \texttt{end\_positions}, ensuring that questions and contexts were properly encoded for fine‑tuning. This preprocessing step is critical, as transformer models rely on tokenized input sequences to learn contextual relationships (Vaswani et al., 2017).

\subsection{Fine-Tuning Setup} 

Models were fine‑tuned using the Hugging Face Transformers library (Wolf et al., 2020) in Google Colaboratory with GPU acceleration. Training parameters included the Adam optimizer (Kingma \& Ba, 2015), cross‑entropy loss for span prediction and fixed values for epochs and batch size. The training process adapted the pre‑trained weights of each model to the domain‑specific dataset, enabling improved performance on university course information.

\subsection{Evaluation} 

Performance was measured using standard extractive QA metrics:

\begin{itemize}
    \item Exact Match (EM) — percentage of predictions that exactly match the ground truth answer.
    \item F1 Score — overlap between predicted and ground truth answers.
\end{itemize}

The fine‑tuned BERT models were evaluated on the validation dataset and compared against their pre‑trained baselines.

\section{Results}

\subsection{Training results} 

We trained each model over several epochs, varying hyper-parameters such as weight decay, learning rate and batch size. For each trial, we recorded both training and validation losses. Across models, a consistent pattern emerged: after approximately four epochs, the validation loss began to increase while the training loss continued to decrease. This divergence indicates the onset of overfitting, suggesting that extended training beyond this point did not yield improvements in generalization performance.

\subsection{Performance results}

\begin{figure}[h]
    \centering
    \includegraphics[width=1\linewidth]{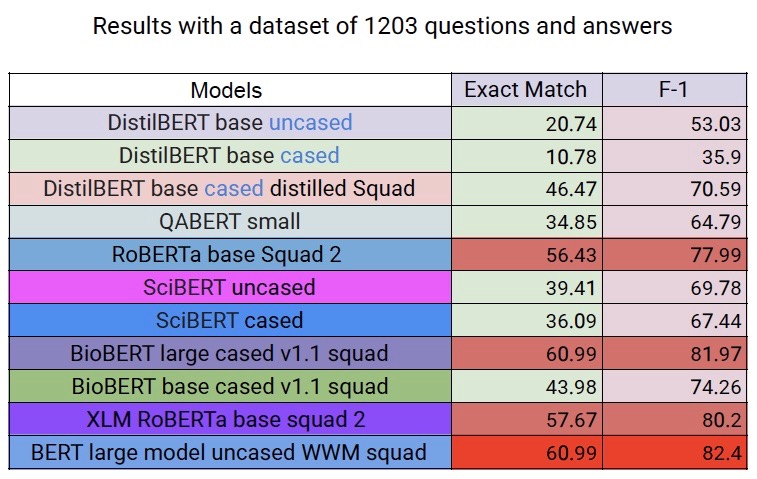}
    \caption{Exact Match and F‑1 scores by extractive question answering models.}
    \label{fig:placeholder}
\end{figure}

The larger models consistently achieved higher scores on both Exact Match (EM) and F1 (Fig. 1). The strongest performance was observed for BERT large uncased WWM SQuAD, followed by BioBERT large cased v1.1 SQuAD, XLM‑RoBERTa base SQuAD 2.0 and RoBERTa base SQuAD 2.0. DistilBERT variants showed comparatively lower EM scores, though all models demonstrated improvements in F1, indicating broader answer coverage even when exact matching was more challenging.

\begin{figure}[h]
    \centering
    \includegraphics[width=1\linewidth]{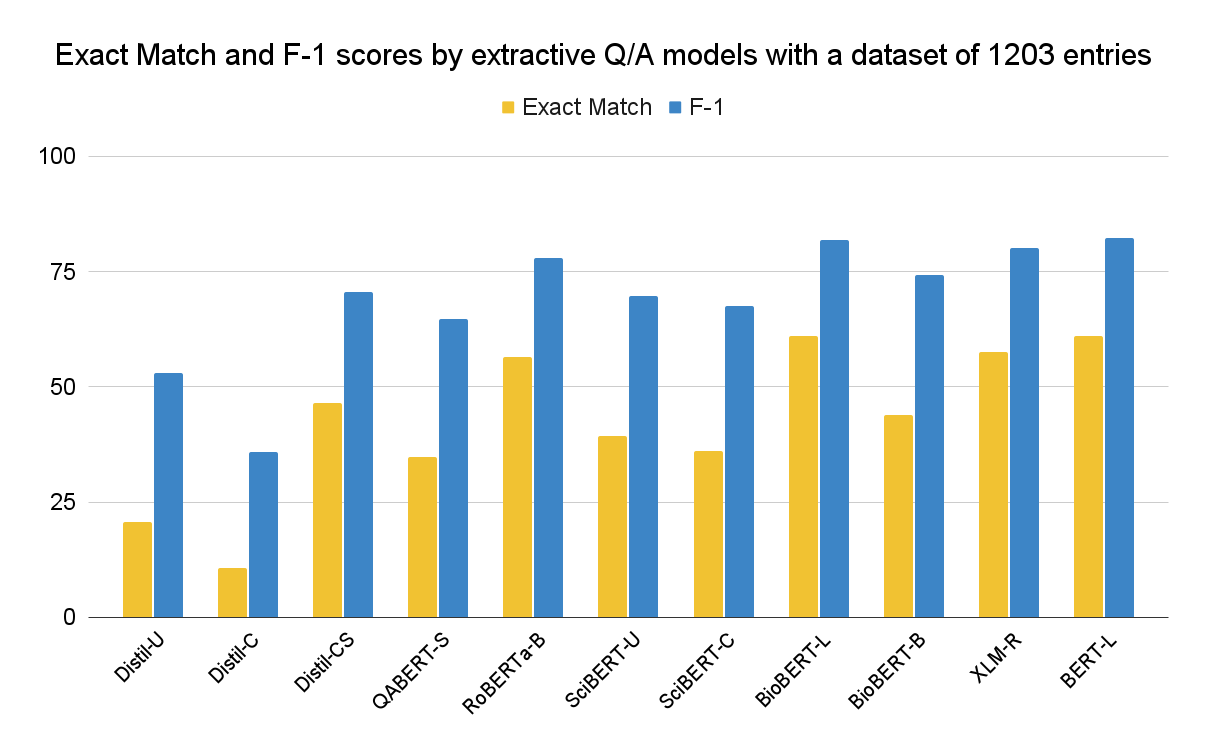}
    \caption{Chart comparing all of the extractive question answering models by Exact Match and F-1 scores. Abbreviations: BERT-L=BERT large uncased WWM SQuAD; BioBERT-L=BioBERT large v1.1 SQuAD; BioBERT-B=BioBERT base v1.1 SQuAD; RoBERTa-B=RoBERTa base SQuAD 2.0; XLM-R=XLM-RoBERTa base SQuAD 2.0; SciBERT-C=SciBERT cased; SciBERT-U=SciBERT uncased; Distil-U=DistilBERT base uncased; Distil-C=DistilBERT base cased; Distil-CS=DistilBERT cased distilled SQuAD.
}
    \label{fig:placeholder}
\end{figure}

RoBERTa base SQuAD 2.0 performed strongly compared to the smaller models, particularly in terms of Exact Match (Fig. 2). The three large models — BERT large uncased WWM SQuAD, BioBERT large cased v1.1 SQuAD and XLM‑RoBERTa base SQuAD 2.0 — showed broadly similar performance, achieving the highest overall scores. This highlights the advantage of larger architectures in maintaining strong accuracy, while RoBERTa base demonstrates competitive results despite being a smaller model.

\subsection{Discussion}

We found that RoBERTa base SQuAD 2.0 performed strongly on the 1,203‑question dataset, achieving an Exact Match score of 56.43 and an F‑1 score of 77.99. SciBERT uncased also performed well, with an Exact Match score of 39.41 and an F‑1 score of 69.78. The BERT large uncased WWM SQuAD model attained the highest scores overall, with an Exact Match of 60.99 and an F‑1 of 82.4, though it showed slight signs of overfitting. Models not already fine-tuned on the original SQuAD dataset generally produced lower scores, suggesting that models already fine-tuned on the original SQuAD provides a clear advantage. Expanding the SQuAD dataset further could potentially improve the Exact Match and F‑1 scores of models such as SciBERT uncased.

\section{CONCLUSION}

We trained and evaluated extractive question answering models on our own dataset in the SQuAD format, consisting of 1,203 question–answer pairs. The models were assessed using Exact Match (EM) and F‑1 scores. Our results showed that models pre‑trained on SQuAD achieved the strongest performance, with BERT large uncased WWM SQuAD attaining the highest EM and F‑1 scores, while RoBERTa base SQuAD 2.0 and SciBERT uncased also demonstrated competitive results. Models not pre‑trained on SQuAD generally produced lower scores, highlighting the importance of domain‑specific pre‑training. A key limitation of this work was the relatively small dataset size, which likely contributed to overfitting and constrained the Exact Match scores. Expanding the dataset further would improve model generalization and performance. For future work, we propose the creation of a large‑scale dataset of university courses and information, with approximately 100,000 entries, similar in scale to the original SQuAD dataset. This dataset could then be used to finetune the models evaluated here, enabling the development of a domain‑specific model tailored for universities. This would represent the first domain‑specific extractive QA resource designed specifically for universities. Institutions could further finetune smaller datasets locally to adapt the model to their specific needs. Such an approach would combine the strengths of large‑scale pre‑training with practical adaptability, supporting more accurate and efficient extractive question answering in academic contexts.

\addtolength{\textheight}{-12cm}   





\end{document}